\documentclass[letterpaper, 10 pt, conference]{ieeeconf}
\IEEEoverridecommandlockouts

\usepackage{geometry}

\usepackage{amsmath,amssymb,amsfonts}
\usepackage{algorithmic}
\usepackage{graphicx}
\usepackage{textcomp}
\usepackage{xcolor}
\usepackage{physics}

\usepackage{epsfig} 
\usepackage{amsopn}
\usepackage{bm} 
\usepackage{subcaption}
\usepackage{float}
\usepackage{algorithm}
\usepackage{dblfloatfix}
\usepackage{gensymb}
\usepackage{array}
\usepackage{mathtools}

\usepackage[
    style=ieee,
    sorting=none
]{biblatex}
\begin{document}

\newgeometry{left = 54pt, top = 54pt, bottom = 54pt , right =54pt}
\title{ \bf Longitudinal Control Volumes: A Novel Centralized Estimation and Control Framework for Distributed Multi-Agent Sorting Systems}

\author{James Maier$^{1}$, 
Prasanna Sriganesh$^{1}$, 
 and  
Matthew Travers$^{1}$
\thanks{ $^{1}$James Maier, Prasanna Sriganesh and Matthew Travers are from The Robotics Institute, Carnegie Mellon University, USA. \texttt{\{jamesmai, pkettava, mtravers\}@andrew.cmu.edu}}
}
\maketitle

\begin{abstract}


 Centralized control of a multi-agent system improves upon distributed control especially when multiple agents share a common task e.g., sorting different materials in a recycling facility. Traditionally, each agent in a sorting facility is tuned individually which leads to suboptimal performance if one agent is less efficient than the others. Centralized control overcomes this bottleneck by leveraging global system state information, but it can be computationally expensive. In this work, we propose a novel framework called Longitudinal Control Volumes (LCV) to model the flow of material in a recycling facility. We then employ a Kalman Filter that incorporates local measurements of materials into a global estimation of the material flow in the system. We utilize a model predictive control algorithm that optimizes the rate of material flow using the global state estimate in real-time. We show that our proposed framework outperforms distributed control methods by 40-100\% in simulation and physical experiments.


\end{abstract}


\section{Introduction}

Centralized control approaches improve task performance over distributed approaches by leveraging information about the system state to coordinate agent actions. In practice, however, difficulties with computational complexity when solving the centralized coordination problem can mean that system designers follow distributed control approaches. One compelling example of a problem domain where practitioners often opt for distributed over centralized control approaches is found in modern facilities for sorting recycled material.

\begin{figure}[!t]
        \centering
        \includegraphics[width = 0.9\linewidth]{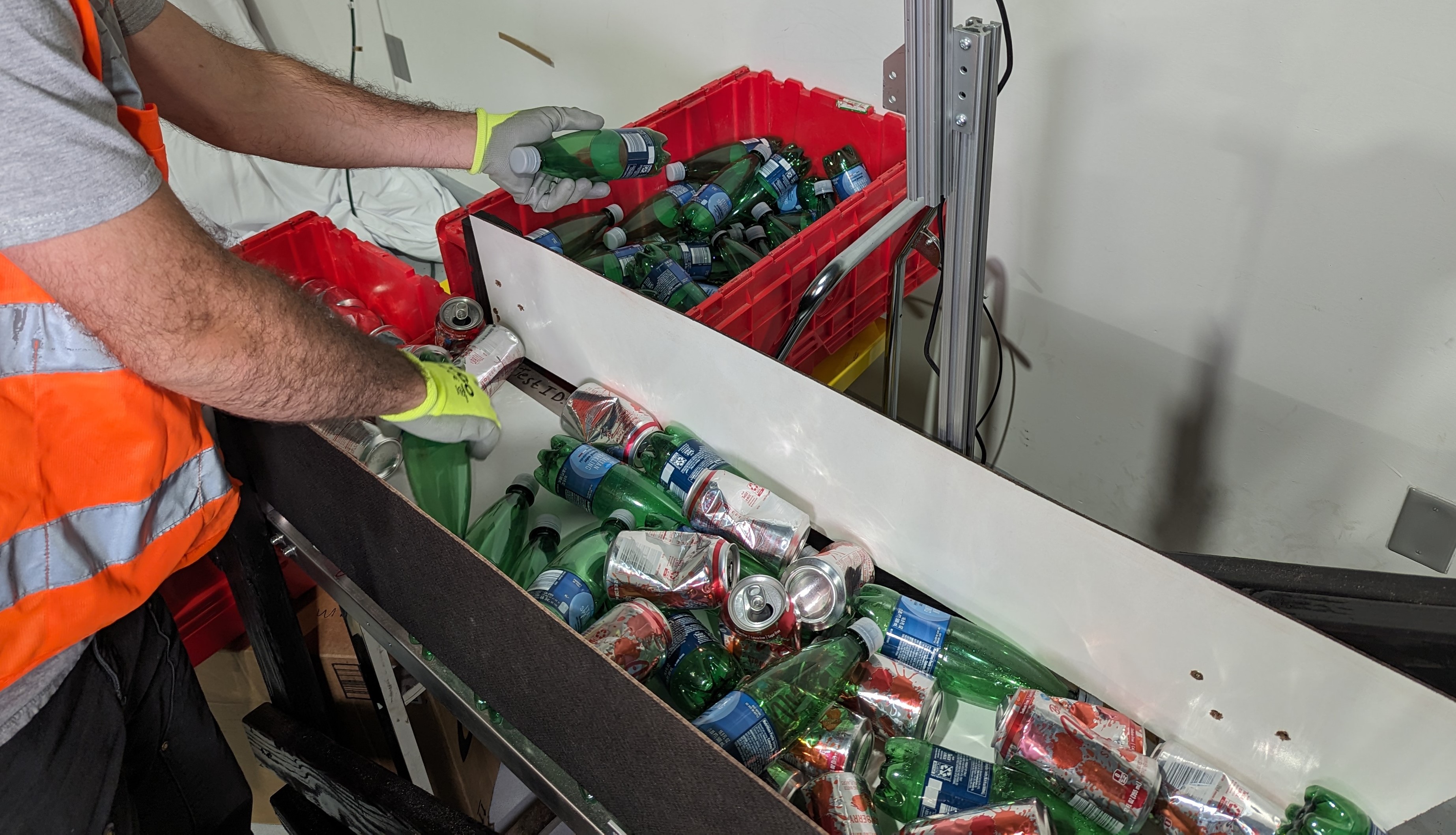}
        \caption{Materials on a conveyor being sorted by a human}
        \label{fig:intro}
        \vspace{-1.5em}
\end{figure}

Recycling sorting facilities generally are comprised of a series of material sorting stations coupled by conveyor belt network. Fig. \ref{fig:intro} shows a simple example of a material sorting operation.  Sorting system operators improve the performance of individual sorting stations by using computer vision systems to detect, classify, and track individual waste items. However, due to the large number of waste items in a sorting system (measured in millions of individual items), these approaches become computationally intractable when they are applied to optimize the system's operation as a whole. 

The control of the flow rate of material through the system is a key aspect of centralized sorting system control. Since the rate at which material is presented to a sorting station directly affects the sort station's ability to remove target material from the stream, a centralized authority can affect the efficacy of sort stations throughout the system by varying flow rates. Dynamically varying the flow rate throughout the system presents a problem for distributed control systems because the system throughput will be limited to that of the slowest station. In contrast, a central authority can avoid bottlenecks by dynamically varying material flow rates in response to varying infeed conditions.

\begin{figure*}[!t]
        \centering
        \includegraphics[width = 0.9\linewidth]{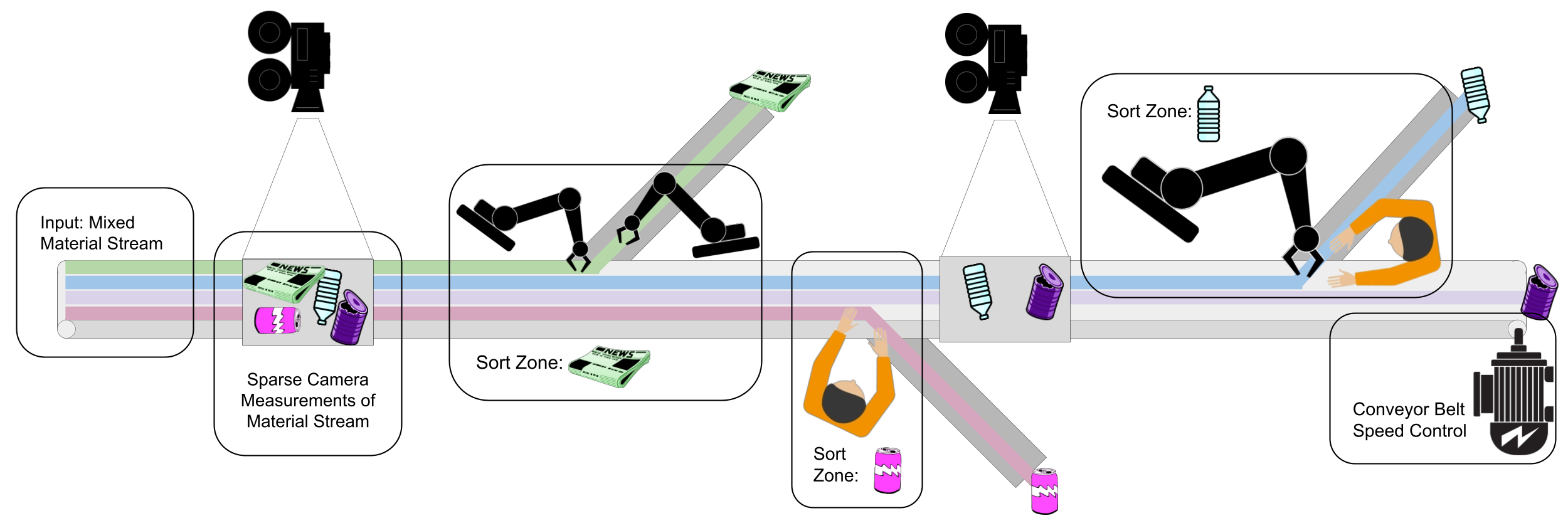}
        \caption{Pictographic representation of multi-agent distributed sorting system with dynamics coupled by a single conveyor belt}
        \label{fig:full_fled}
        \vspace{-1em}
\end{figure*}

In this work, we present a novel modeling framework that divides the entire conveyor into smaller Longitudinal Control Volumes (LCV) which enables a low-dimensional representation and prediction of material flows throughout the sorting facility. We present a method for converting detections of individual waste items from images into sparse measurements in our state space. We use these measurement and prediction framework with a Kalman Filter to compute a global estimate of the system state.  

This modeling framework in is used conjunction with the global estimation result in a model predictive control (MPC) module that optimizes the material flow to maximize the overall value generated for successful sorting. The controller reacts to the incoming material over a receding time horizon and changes the speed of the conveyor to allow for efficient sorting of materials. 

We demonstrate the efficacy of our centralized estimation and control approach in simulation and on a physical laboratory sorting system.  We show improvements on the order of 40-100\% with respect to constant belt speed approaches. We later discuss applications of our framework to diverse multi-agent systems including autonomous mobile robots and over-actuated control systems.

\section{Related Work}\label{relatedwork}

In order to estimate performance of the core sorting system in a material recovery facility (MRF), the authors in \cite{tr2021a}\cite{wolf2011}\cite{testa2015} develop tools to estimate sorting process efficiency based on historical data on material flow compositions.  These tools leverage long-term historical data and are used to calculate steady-state optimal material flow rates for a given system.  In this system-level offline modeling approach, a network flow diagram for the MRF is created by approximating the dynamics of mixed material flows through sorting machines and connecting them via a directed graph.  A system of linear equations is then constructed and solved to calculate material stream compositions throughout the system at steady state \cite{gutowski}\cite{wolf2011}. 

The work presented in \cite{testa2015}\cite{ip2018}\cite{kleinhans} expands on the foundational work by \cite{gutowski} and \cite{wolf2011} by estimating the percentage of material in the main stream that a single sort station can remove.  They then set up a centralized optimization problem to find nominal steady-state optimal material flow rates. These offline modeling and optimization techniques represent a method for achieving global system coordination. However, they do not capture dynamics associated with material flow characteristics changing in time.  They also rely on assumption of static sorting parameters which are often not representative of real sorting operations \cite{caputo}\cite{velis}. 

A second line of research involves distributed control techniques which can run in real-time. In these distributed control approaches, individual sorting machines leverage local system information to improve their performance.  A literature review of these approaches for distributed optimization of sorting systems was conducted by \cite{kroell22}. 

Of particular relevance to the content of this work, \cite{pfaff17}\cite{georg22}\cite{pfaff19} leverage techniques from computer vision and machine learning along with Kalman Filtering and multi-target tracking.  They track each object entering a sorting machine and optimize that sorting machine's operation around real-time flow characteristics.  These approaches show promising improvements to the operations of individual sorting machines, but they have not been applied to solving the global coordination problem.  We hypothesize that the computational complexity of simultaneously tracking millions of individual pieces of recycled material presents an obstacle to use in centralized control methods. 

Furthermore, as noted in \cite{raymond2017} and \cite{tr2021a}, real MRF operations are much less efficient than would be expected.  The authors of \cite{tr2022} investigate the discrepancy between theoretically expected sorting operations and real MRF sorting data and conclude that the high degree of variability in the material stream means that solutions calculated with offline centralized optimization systems do not translate well to real operation. The authors in \cite{curtis20} study the variability in the input material stream and provide statistics on flow variability from real MRF operations, and they establish that online control techniques are necessary to cope for the variations in the input material flows.

In this work, we draw particular inspiration from \cite{tr2022} and \cite{curtis20} who establish that MRF sorting performance suffers in the face of variable input material flows.  We develop a state modeling approach which follows ideologically from the systems described in \cite{wolf2011} and \cite{testa2015}, but which provides a measure of system state with enough fidelity for real-time dynamic control.  We use this state estimation result as a basis for real-time centralized control of material flows.

\section{Modeling Material Flow Compositions}

In this section, we present a method to represent the current state of a sorting system using a framework called Longitudinal Control Volumes. We then describe a model that can track the material composition flow as it moves across the conveyor.

\subsection{Longitudinal Control Volumes (LCV)}

\begin{figure}[h!]
\vspace{-1em}
\centerline{\includegraphics[width = 0.7\linewidth]{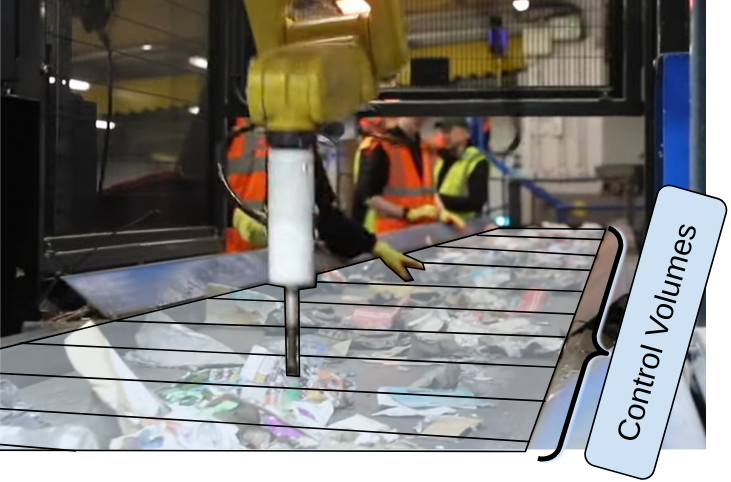}}
\caption{Control volumes overlaid on conveyor belt serving a team of human and robot sorters}
\label{cont_vol}
\vspace{-1em}
\end{figure}

The state representation of a sorting system needs to provide information about the composition of the material flow as it evolves along the conveyor. To compute this, we first longitudinally split the sorting system into small control volumes arranged sequentially along the direction of travel of the belt (Fig. \ref{cont_vol}). The size of these control volumes is tuned based on the spatial resolution of the perception system and the computational load of having a high number of control volumes. The amount of material in the control volume forms the basis for constructing the state vector.  

We construct the state vector by first concatenating smaller vectors $\vectorbold{x^{i}_k}$,  for every material~$i$. Each vector element in $\vectorbold{x^{i}_k}$ corresponds to a control volume and indicates the quantity of a material occupying that control volume at timestep $k$. Lastly, the current speed of the conveyor belt, $r_k$ is added to the end of the concatenated vector which forms the full system state. The total amount of a given material in the entire system can by computed by summing the elements of the state vector corresponding to that material. Given a total of $m$ control volumes and $n$ different material types to be sorted, the state vector is of the size $nm + 1$ and can be written as follows
\begin{align}
    & \vectorbold{X_k} = [\vectorbold{x^1_k}, \vectorbold{x^2_k}, \dots , \vectorbold{x^n_k}, r_k]^T  \\
    & \textit{Total amount of material $i$ in the system} = \sum \vectorbold{x^i_k} \nonumber 
    \vspace{-2em}
\end{align}
   
\subsection{State Transition Model} \label{sec_3b}

The motion of the conveyor belt and the sorting process change the composition of material in the control volumes. To predict this change in the system state, we decompose the state transition matrix to model both the dynamics associated with the conveyor motion ($\mathcal{L}(r_k)$) and the dynamics associated with the sorting process ($\mathcal{F}(\vectorbold{X_k})$) to model a linear approximation of these processes. The control input to the system is the change in conveyor speed $\Delta r$. This makes the matrix $B$ a $mn + 1$ long zero vector with the last element equal to 1 as the control input affects only the last element of the state.  This state transition model can be expressed as follows.
\begin{align}
    \vectorbold{X_{k+1}} &=  A (\vectorbold{X_k})\vectorbold{X_k} + Bu_k  \nonumber \\
     &=  \mathcal{F}(\vectorbold{X_k}) \mathcal{L}(r_k) \vectorbold{X_k}  + B \Delta r
\end{align}

The motion matrix $\mathcal{L}(r_k)$ models the motion of material across different control volumes in the system carried by conveyor belts. This is dependent on the current speed of the conveyor belt. To construct this matrix for any arbitrary belt speed, we compute these motion matrices for integral belt speeds and perform linear interpolation between these integral speed motion matrices. Given an integer belt speed $\hat{r_k}$, we can construct a motion matrix $\mathcal{L}(\hat{r_k})$ as follows,

\begin{align*}
 \mathcal{L}(\hat{r_k}) &= \begin{bmatrix}
                    L(\hat{r_k}) & 0_{m\times m} & \dots & 0_{m\times m} & 0 \\
                    0_{m\times m} &  L(\hat{r_k}) & \dots & 0_{m\times m} & 0 \\
                    \dots  &  \dots &  \dots & \dots  & .  \\
                    0_{m\times m} & 0_{m\times m} & \dots & L(\hat{r_k}) &  0  \\
                    0 & 0 & \dots & 0 & 1 
                \end{bmatrix} \\
\textit{where } & \textit{$L(\hat{r_k})$ is a $m \times m$ modified shift matrix given by} \\
 L(\hat{r_k})_{ij} &= \delta_{i,j+\hat{r_k}} \\
 \textit{ where } & \textit{ $\delta_{i,j}$ is the Kronecker delta}
\end{align*}

This motion matrix can only be defined for an integral belt speed due to the Kronecker delta. Hence we do a linear interpolation of the lower and upper integral motion matrices to compute the motion matrix for an arbitrary belt speed. Given a belt speed $r_k$, we can compute the motion matrix $M(r_k)$ as follows, 
\begin{align*}
    \mathcal{L}(r_k) &=  \frac{ \lceil r_k \rceil - r_k}{ \lceil r_k \rceil - \lfloor r_k \rfloor} \mathcal{L}(\lfloor r_k \rfloor) +  \frac{r_k - \lfloor r_k \rfloor}{ \lceil r_k \rceil - \lfloor r_k \rfloor} \mathcal{L}(\lceil r_k \rceil )\\
    \textit{where } & \lceil x \rceil = ceil(x) \textit{ and } \lfloor x \rfloor = floor(x)
\end{align*}

We model the dynamics associated with removing material at sorting stations using the sort matrix $\mathcal{F}(\vectorbold{X_k})$. As a sorting station only spans a few control volumes in the entire system, it can sort a given material at a fixed upper limit. To capture this behavior, we define a separation parameter $p^{i}_{j}$ for each sort station \cite{tr2021b}\cite{gutowski}. This separation parameter is defined per material for every control volume based on the maximum number of materials that can be picked up by a sort station.

If $\boldsymbol{\alpha^i}$ represents the subset of control volumes spanned by a sort station and $\gamma_i$ represents the maximum amount of material that can picked up in a single timestep, we can compute the separation parameters based on ratio of maximum pick rate and amount of material present on the control volumes of a sort station ($\eta$). For example,  when the amount of material is less than the maximum material that can be picked up, $\eta^i$ is greater than $1$ which makes the separation parameter 0 to indicate that all the material has been picked up.
\begin{align*}
    \eta^i &= \frac{\gamma^i}{\sum\limits_{\boldsymbol{\alpha^i}} \vectorbold{x^i}} & p^i_j &= \begin{cases}
            1,  &  \textit{if } j \notin \boldsymbol{\alpha^i} \\
        max(0, 1 - \eta^i),    & \textit{otherwise} 
    \end{cases} 
\end{align*}

The sorting $ \mathcal{F}(\vectorbold{X_k})$ matrix consists of smaller diagonal matrices $F(\vectorbold{x^i_k})$ for that capture the sorting process for every material in each control volumes. The matrix $F(\vectorbold{x^i_k})$ is a $m \times m$ diagonal matrix with the diagonal elements equal to the separation parameter $p^i_j$. We can write this as follows:

\begin{align*}
    \mathcal{F}(\vectorbold{X_k}) &= \begin{bmatrix}
                    F(\vectorbold{x^{1}_k}) & 0_{m\times m} & \dots & 0_{m\times m} & 0 \\
                    0_{m\times m} &  F(\vectorbold{x^{2}_k}) & \dots & 0_{m\times m} & 0 \\
                    \dots  &  \dots &  \dots & \dots  & .  \\
                    0_{m\times m} & 0_{m\times m} & \dots & F(\vectorbold{x^{n}_k}) &  0  \\
                    0 & 0 & \dots & 0 & 1 
                \end{bmatrix}\\
            F(\vectorbold{x^i_k}) &= \begin{bmatrix}
                p^i_1 & 0 & \dots & 0 \\
                0 & p^i_2 & \dots & 0 \\
                . & . & \dots & . \\
                0 & 0 & \dots & p^i_m 
            \end{bmatrix}
\end{align*}

\section{Methodology}

\begin{figure}[!h]
        \centering
        \vspace{-1em}
        \includegraphics[width = 0.7\linewidth]{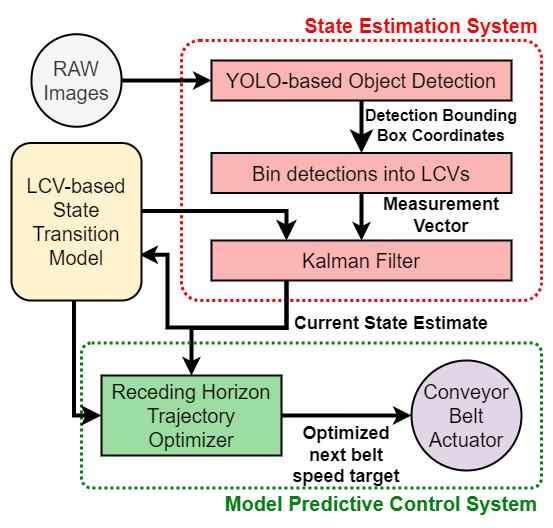}
        \caption{A block diagram of the proposed system to optimize belt speeds for sorting}
        \label{fig:block_diagram}
        \vspace{-1.5em}
\end{figure}

We create a system to make an optimal tradeoff between sorting task completion and system throughput in the context of a recycling sorting operation. As the material flows through the conveyor, there is uncertainty in the state induced by the sorting process and material moving relative to the conveyor belt. To incorporate this uncertainty, we present a framework to use a computer vision system to detect materials and estimate the system's state using an Kalman Filter based on the LCV state transition model. We then use this estimated state as input for a centralized model predictive control module, which maximizes the difference between the reward associated with successful sorting and the cost associated with failing to sort potentially valuable material. The controller outputs an optimized conveyor belt speed that maximizes the system's throughput while maintaining high rewards for sorting. A block diagram of this system is shown in Fig. \ref{fig:block_diagram}.

\subsection{State Estimation}

In order to incorporate real-time data into our state estimate, we devise a computer vision system to detect individual objects on a subset of control volumes. An Intel Realsense D435 RGB camera provides top-down images to a YOLO-based object detection system that classifies different materials and returns bounding boxes that describe the location and type of individual items \cite{yolo}. The detection from every frame is incorporated as a measurement vector. 

To compute the measurement vector, we project the visible control volumes into the camera image and add the counts of detected objects into the control volumes associated with their locations. If a detected object spans multiple control volumes, we add portions of the detected object to each control volume spanned by the object. If $\lambda$ is the number of control volumes in the camera's view and $W^i$ is the set of detections of material $i$, we can compute the measurement vector $\vectorbold{z_k}$ as follows,
\begin{align*}
    \vectorbold{z_k} &= \begin{bmatrix}
        \Bar{\zeta^0} & \Bar{\zeta^1} & \dots & \Bar{\zeta^n}
    \end{bmatrix}  & \Bar{\zeta^i} = \begin{bmatrix}
        \beta^i_0 & \beta^i_1 & \dots & \beta^i_\lambda
    \end{bmatrix}
\end{align*}
\begin{align*}
    \beta^i_j &= \sum_{w \in W^i} \frac{ \begin{matrix} \textit{Area of bounding box for detection } w \\ \textit{ spanning control volume }j \end{matrix}} { \begin{matrix}
        \textit{Total area of bounding box} \\ \textit{ for detection } w
    \end{matrix}} 
\end{align*}

\begin{figure}[t!]
    \centerline{\includegraphics[width=0.7\linewidth]{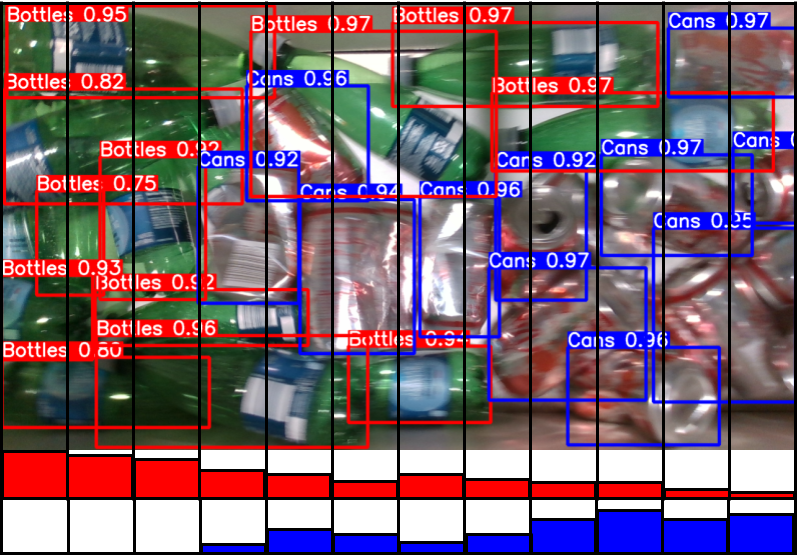}}
    \caption{Control volume boundaries overlaid on image with bounding box detections from object detection system.  The red and blue bars at the bottom help illustrate binning for different materials and the measurement vector generated from this input image.}
    \label{yolo_vols}
    \vspace{-1.5em}
\end{figure}

This process is shown in Fig. \ref{yolo_vols}. The measurement vector is of the size $n\lambda$ as it comprises the control volumes visible to the camera. We also associate a measurement covariance matrix to incorporate the noise inherent to the object detection system.

We adopt a Kalman Filter (KF) that can estimate the entire state of the system using the state transition model and the measurements coming from the computer vision system \cite{kf}. The uncertainty induced by the sorting process and movement of the conveyor belt is modeled as the process noise for the KF. This filter enables the prediction of the full-state with an associated covariance that represents the uncertainty in the predicted state.

\subsection{Model Predictive Control}

We setup a Model Predictive Control (MPC) problem which uses the state-transition model described in \ref{sec_3b} and the estimated state to optimize sorting operations and control the conveyor belt speed in real-time. We choose to maximize an objective function that can maximize the value of sorted material which is obtained using a Value matrix $V$ and an Opportunity Cost matrix $O$. The Value matrix maps the price of a material $\rho_i$ onto the sort matrix($F^i(\vectorbold{x_k}))$. The Opportunity Cost matrix encodes the material that we fail to sort and are forced to sell for a lower price as mixed material. These matrices are computed as shown below.
\begin{align*}
    V &= \begin{bmatrix}
        \Bar{v}^0 & \Bar{v}^1 & \dots & \Bar{v}_{n} & 0 
    \end{bmatrix} & \Bar{v}^i &= \rho^i F(\vectorbold{x^{i}_k}) \vectorbold{1_m} \\
    O &= \begin{bmatrix}
         \Bar{o}^0 & \Bar{o}^1 & \dots & \Bar{o}_{n} & 0 
        \end{bmatrix} & \Bar{o}^i &= \rho^i (I - F(\vectorbold{x^{i}_k})) \vectorbold{1_m} \\
       &&&& \llap{where $\vectorbold{1_m}$ is a 1-vector of size $m \times 1$} 
\end{align*}
 For the optimization, we set the horizon to the time it would take material to fully transit the system with all belts at their slowest speed. Eq. \ref{obj_fcn} describes our objective function that seeks to maximize the difference between the total sale price of material sorted over the horizon and the value lost by selling a material with contamination.
\begin{align} 
        \max_{X_{1:T}} \quad & \sum_{l=1}^{T} [V - O] \vectorbold{X}_l  \label{obj_fcn} \\
            \textit{s.t.}  \nonumber \\
             X_{l+1}  = \mathcal{F}(\vectorbold{X_l}) &\mathcal{L}(r_l) X_l + Bu_l \quad \textit{for } l = 1,\ldots, T-1 \nonumber \\
            \quad & u_{min} \leq  u_{l} \leq u_{max} \nonumber
\end{align}
 We set up a quasi-Newton solver with a back-tracking line search to maximize our objective function. We compute the gradients by calculating finite-difference derivatives for each control variable, and use the gradient for to pick the update direction. The update step is then scaled using a BFGS approximation of the inverse Hessian matrix in the traditional Newton update equation \cite{broyden}\cite{fletcher}\cite{goldfarb}\cite{shanno}. The optimization is terminated based on fulfillment of the Armijo Condition\cite{armijo}. The resulting optimized control input (conveyor belt speed) corresponding to the next timestep is then applied to the system. This process is repeated at every timestep.

\section{Experiments}

We conduct experiments in simulation and on a physical conveyor belt sorting system and evaluate our system's performance by comparing it to that of a baseline approach. For the baseline, we run the conveyor at a constant speed, similar to approaches currently used in industrial MRFs. This optimal constant speed is estimated based on the historical average material flow through the system. We compare the total value generated by this baseline to the value generated by running the system at the optimized variable belt speed generated by the MPC module. 
\subsection{Simulation-based evaluation of MPC performance}

We create a simulated environment where infeed material flow is procedurally generated in a manner representative of actual material flows in an MRF according to material variability statistics described in \cite{curtis20}. The LCV-based state transition model described previously is used to forward-simulate the sorting operation. We generate 30 runs that each represent an hour of continuous operation. We take the average optimized belt speed produced by the MPC module for each run and repeat the experiment with a constant belt speed equal to that average optimized speed. We compare the final accumulated value at the end of each optimized run with the accumulated value from its corresponding constant speed run. We repeat this process for sorting systems with 2, 3, and 4 distinct materials to be sorted.

\subsection{Full-system Experiments on Physical Sorting Equipment}
\begin{figure}[b!]
    \centering
    \vspace{-1.0em}
    \includegraphics[width = 0.75\linewidth]{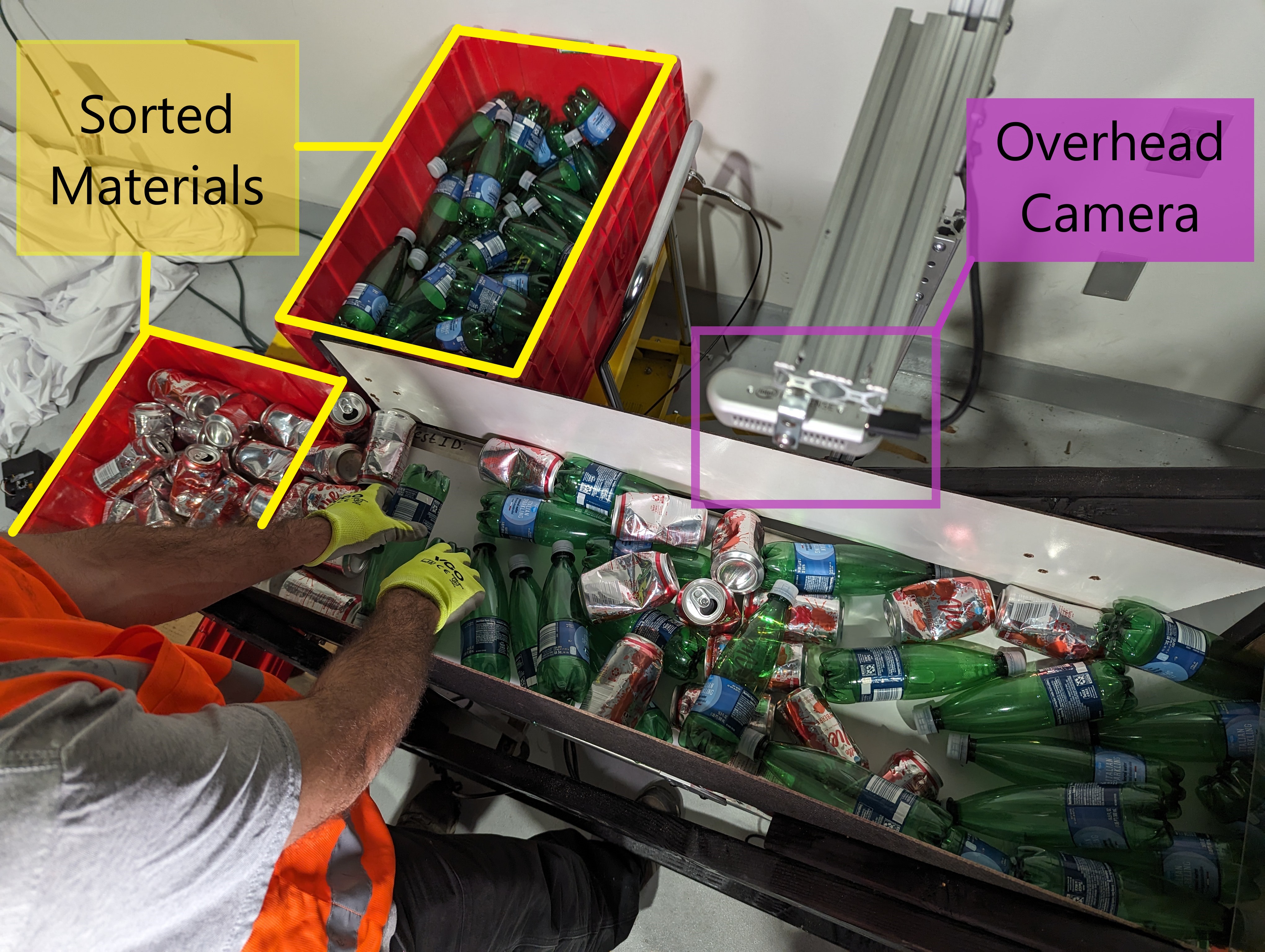}
    \caption{Physical conveyor belt experimental setup showing the RealSense D435 overhead camera, human sorter and the sorted materials}
    \label{fig:expsetup}
\end{figure}

We also conducted experiments on a physical conveyor belt sorting station to evaluate the performance of our end-to-end system. Different material streams can be fed to the conveyor belt using a hopper on one end. A computer with a ROS framework captured overhead images from the Intel RealSense D435 camera, executed the state estimation and MPC modules in real-time, and controlled the speed of the conveyor belt using a servo motor. At the end of this conveyor, a human sorter was tasked with removing one material from the flow. This setup is shown in the Fig.~\ref{fig:expsetup}. 

In these physical experiments, we use two materials: empty green bottles and empty red cans. In order to investigate the ability of our framework to react to changing infeed conditions during these short runs, we loaded the hopper with varying distributions of the material flows as the MPC module optimized the conveyor belt speed in real-time. For each of these runs, we computed the value generated for sorting and compared this to the value generated using a constant speed baseline experiment. The constant speed was equal to the average of the optimized speeds generated by the MPC module.
 
\begin{figure}[!b]
        \centering
        \vspace{-1.5em}
        \includegraphics[width = 0.9\linewidth]{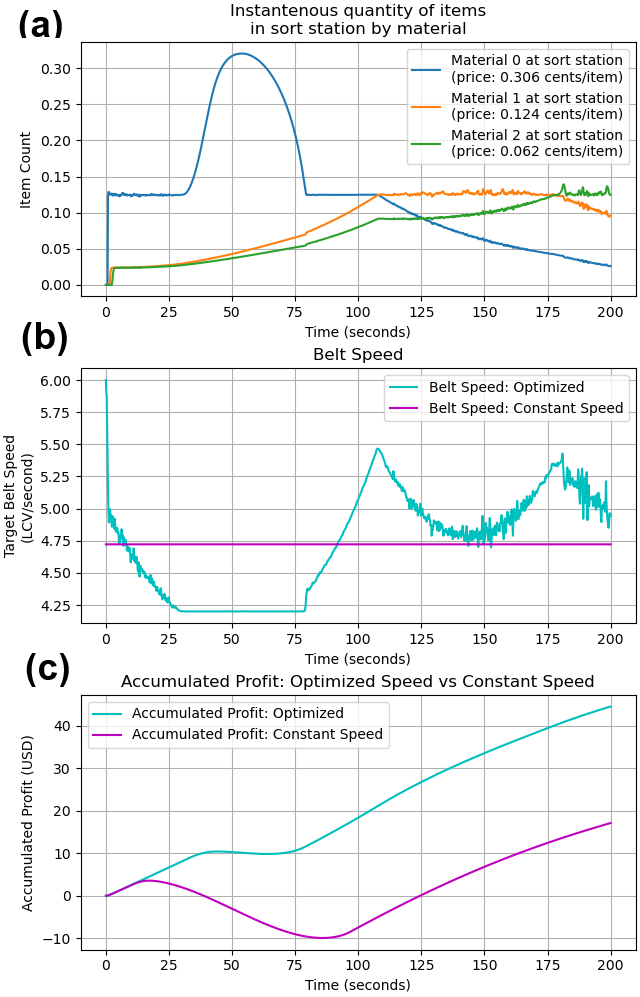}
        \caption{Result of a simulated 3-material sorting run. (a) Plot showing the instantaneous amount of material passing through a sort station (b) Plot showing the output belt speeds for the MPC and the baseline. (c) Plot showing the accumulated value as a result of successfully sorting.}
        \label{fig:3mat}
\end{figure}

\section{Results}

The results of our proposed framework for a three-material sorting simulation is shown in the Fig. \ref{fig:3mat}. We compare the total profit generated by our method to that of the constant speed baseline model. Fig. \ref{fig:3mat}(a) shows the instantaneous distribution of the materials in their respective sort stations, along with their price. Fig. \ref{fig:3mat}(b) shows shows the target conveyor belt speed by the MPC module (cyan line) and the target belt speed by the constant speed baseline (magenta line). Fig. \ref{fig:3mat}(c) shows the total profits accumulated for successfully sorted materials for both versions. 

\begin{figure}[!t]
        \centering
        \includegraphics[width = 0.9\linewidth]{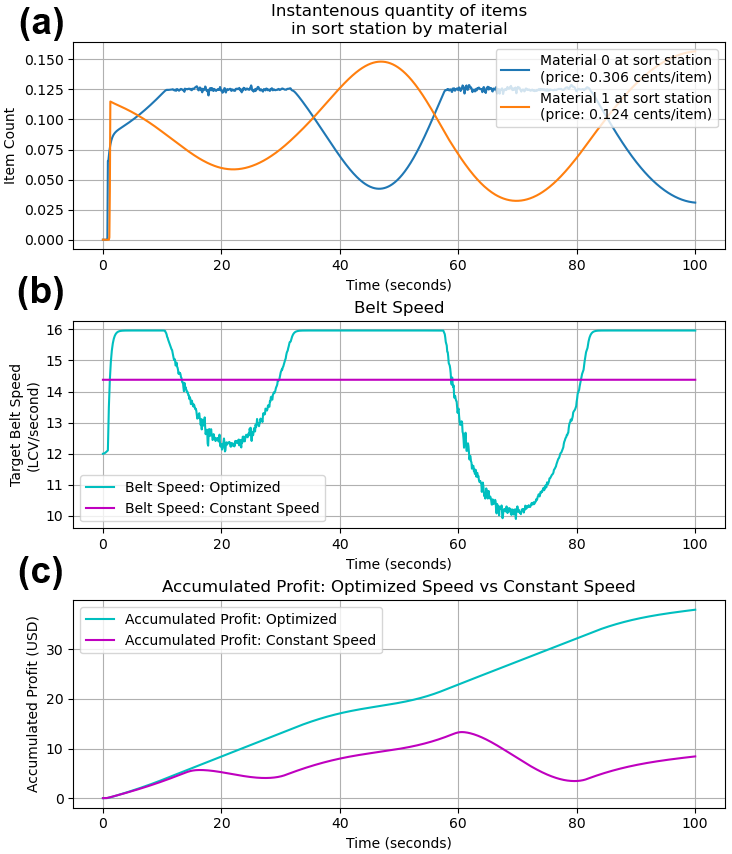}
        \caption{Result of a 2-material sorting run. (a) Plot showing the instantaneous amount of material passing through a sort station (b) Plot showing the output belt speeds for the MPC and the baseline. (c) Plot showing the accumulated value as a result of successfully sorting.}
        \label{fig:2mat}
        \vspace{-1.0em}
\end{figure}

From the plots shown in the figures \ref{fig:3mat}(a) and \ref{fig:3mat}(b), it is evident that the MPC module is reacting to the amount of material coming into the system. As more Material~0 (which has a higher price) enters the system, the controller slows down the conveyor to allow the sorters to pick up more material, increasing the profits accumulated. For the same case, since the constant speed model does not change the belt speeds, the plant incurs a loss as the sorters fail to sort this high-value material, and the impure mixture is worth much less. We see the same behavior even in the 2-material runs as shown in the Fig. \ref{fig:2mat}.

\begin{figure}[!t]
        \centering
        \setlength{\tabcolsep}{2.5pt}
        \begin{tabular}{c c}
           \subfloat[Simulated material flow]{\includegraphics[width = 0.55\linewidth]{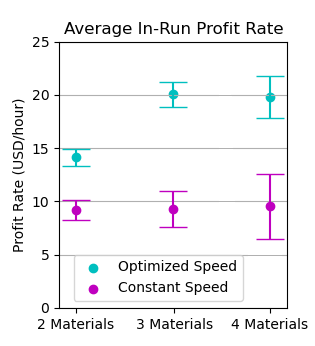}}   &  
           \subfloat[Physical experiment with 2 materials]{\includegraphics[width = 0.35\linewidth]{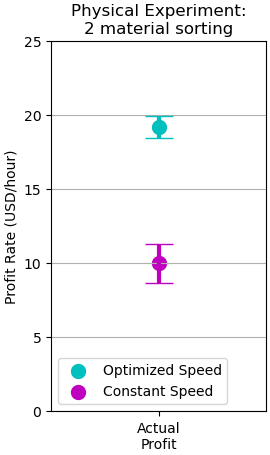}} 
        \end{tabular}
        \caption{Result plot showing profit rate for optimized speed vs constant speed control} 
        \label{fig:errorbars}
        \vspace{-1.5em}
\end{figure}

We notice similar trends across all the different runs for varying amounts of materials in the system. When different materials are distributed evenly, the constant speed baseline and our proposed framework perform similarly. In contrast, the profit rate of our method goes up quickly when the distribution varies quickly. This was true across all the 90 runs we conducted in simulation. Fig.~\ref{fig:errorbars}(a) shows the average profit rate along with the variance in profit rate for the proposed method and the constant speed baseline for varying number of simulated material runs. In the 2-material case, our optimized speeds improved the margin by around 40 percent while we see around two times the profit in the 3-material and 4-material cases. In the physical experiments with 2-materials, the optimized speed run improves the profit margins by two times as shown in Fig. \ref{fig:errorbars}(b).

\section{Conclusion and Future Work}
In the paper, we presented a novel framework that modeled the movement of material along a conveyor belt in a sorting system. We presented an end-to-end system that successfully detected and estimated these material flows while optimizing the conveyor belt speed for maximizing value in the sorting process. We conducted experiments both in simulation and on a physical conveyor belt with varying numbers of materials and compositions. We achieved around 40 to 100\% increase in the profit rates using the proposed model predictive controller that accounted for the entire state of the system. 

Although the presented work shows dramatic improvements to process profitability, we identify several potential avenues for further research. As the MPC module does not leverage the uncertainty information from the state estimation module, we intend to explore incorporating the uncertainty in the optimization process along with trying to model an objective function derived from operations research. We would like to investigate learning-based approaches to directly infer measurement vectors from camera images. Further, we are in the process of scaling our experiments at an operational MRF sorting facility.

Beyond applications in conveyor based sorting, we firmly believe that use of control volumes is more broadly applicable to systems in settings that have a team of agents team collaborating on a task.  Specifically, we intend to apply this approach to problems associated with a team of autonomous mobile robots exploring an unknown area \cite{bagree2022high}\cite{sriganesh2023fast}\cite{scherer2022resilient}.  We also intend to investigate this approach in the control of micro-robot swarms for medical applications \cite{benjaminson2020steering}\cite{benjaminson2023buoyant}.

\printbibliography

\end{document}